\documentclass[lettersize,journal]{IEEEtran}
\usepackage{amsmath,amsfonts}
\usepackage{algorithmic}
\usepackage{algorithm}
\usepackage{array}
\usepackage[caption=false,font=normalsize,labelfont=sf,textfont=sf]{subfig}
\usepackage{textcomp}
\usepackage{stfloats}
\usepackage{url}
\usepackage{verbatim}
\usepackage{graphicx}
\usepackage{cite}
\usepackage{makecell}
\usepackage{placeins} 
\usepackage{longtable}
\usepackage{comment}
\usepackage{multirow}
\usepackage{hyperref}
\usepackage{orcidlink}
\usepackage{xcolor}
\usepackage{multirow}
\usepackage{adjustbox} 
\usepackage{hyperref}
\usepackage{booktabs}
\usepackage{hyperref}
\usepackage{pifont}

\newcommand{\cmark}{\ding{51}}
\newcommand{\xmark}{\ding{55}}
\begin{document}

\title{RAGMesh with FaME-G2E: Long-Form Text-Driven 3D Face Generation and Editing}

\author{Hao Li, Ju Dai$^{*}$, Feng Zhou, Mengting Shi, Haofei Wang, Zhen Song, Wei Zhou,~\IEEEmembership{Senior Member~IEEE}, Lei Li, Junjun Pan$^{*}$
\thanks{
Corresponding authors: Ju Dai (daij@pcl.ac.cn) and Junjun Pan (pan\_junjun@buaa.edu.cn). \\
Hao Li and Junjun Pan are with the State Key Laboratory of Virtual Reality Technology and Systems, Beihang University, Beijing, China.  
Hao Li, Ju Dai, Haofei Wang, and Zhen Song are with Pengcheng Laboratory, Shenzhen, China. Mengting Shi is with Shanxi University of Finance and Economics, Shanxi, China. Feng Zhou is with North China University of Technology, Beijing, China. Wei Zhou is with Cardiff University, Cardiff, UK, Lei Li is with Beijing Institute of Technology, Beijing, China. 
}}



\maketitle

\begin{abstract}
Text-driven 3D face generation and editing remains challenging due to the difficulty of translating long-form descriptions into fine-grained facial geometry. Existing methods primarily align global textual semantics with facial structures but often struggle to capture subtle local deformations, such as eyebrow tension, cheek contraction, and asymmetric mouth motions, resulting in limited geometric fidelity and editing precision. To facilitate fine-grained text-driven facial modeling, we first construct FaME-G2E, a large-scale multimodal dataset containing detailed text--mesh annotations and paired text--blendshape samples for unified 3D facial generation and editing. Based on this dataset, we propose RAGMesh, a retrieval-augmented framework that leverages text-correlated geometric priors to improve high-fidelity facial synthesis and editing. Specifically, the Multi-Scale Retrieval Fusion (MSRF) module retrieves semantically consistent global and regional facial priors and fuses them in the blendshape space, suppressing conflicting local deformations while preserving coherent deformation patterns. Furthermore, we introduce Adaptive RAG-guided Supervision (AdaRAGS), a region-aware constraint that explicitly aligns textual semantics with corresponding facial regions, enhancing regional controllability and editing accuracy. Extensive experiments on FaME-G2E demonstrate that RAGMesh achieves superior performance over state-of-the-art methods in local geometric accuracy, text-guided controllability, regional editing precision, and inference efficiency. Video demo is available at \url{https://youtu.be/Yr0_XkpWcNk}, and the source code and dataset will be released upon paper acceptance.
\end{abstract}




\section{Introduction}
\IEEEPARstart{T}ext-driven 3D facial generation and editing aim to achieve controllable creation and manipulation of human faces via natural language descriptions~\cite{3D-TOGO,Magic3D,LiuWQF22,abs-2509-02466,dey2022generating}. These tasks enable intuitive and flexible expression modeling for virtual humans and character animation~\cite{liang2024skull,song2024expressive,zhang2023meshwgan,zhuang2026talkingeyes,ling2022semantically}. However, existing models struggle to align complex and structured long textual descriptions with local facial geometry, often confusing direction-sensitive regions such as the left and right eyes or mouth corners, resulting in ambiguous or mirrored expressions that deviate from the intended textual semantics~\cite{DreamFace,ICE,FaceG2E,wood20223d,YouwangKO22}. 

The generation results of early end-to-end modeling frameworks often focus on text tokens with the most substantial gradients while neglecting region-specific or subtle descriptive cues~\cite{CLIP, Wang0ZGBBS00CG23, ClipFace, GeXJPTLDJZ25}. Methods optimized via gradient supervision derived from CLIP-based 2D priors (e.g., SDS \cite{FaceG2E}) effectively capture global semantics, but their performance is constrained by the resolution and expressiveness of the pretrained feature space, making it difficult to handle fine-grained descriptions in long textual inputs~\cite{DreamFace,FaceG2E,abs-2507-05256,ZhengLLXNH25}. Recently, LLM-based parsing methods, such as ICE~\cite{ICE}, map parsed textual semantics onto predefined BlendShape weights. Although these approaches provide interpretability and controllability, their overall coherence remains suboptimal. In summary, existing methods~\cite{HanC0ZDS0W23,TADA,FaceG2E} struggle to achieve precise alignment between complex, structured long textual descriptions and localized facial geometric deformations.

Recently, several works~\cite{SeoHJKKLK24,SDS1} such as RetDream~\cite{RetDream} for text-to-3D reconstruction and AMD~\cite{AMD} for text-to-motion generation have demonstrated that introducing external priors through retrieval-augmented generation (RAG) effectively enhances semantic richness and diversity across various AIGC tasks~\cite{Komeili0W22,Zhou0XJN23,0001NQXBA23,YangCZ23}. By retrieving task-relevant knowledge or examples, RAG can provide additional prior knowledge, enhance contextual grounding and enhance the fidelity of generated content. Inspired by these successes, we argue that retrieving semantically similar local geometries from the retrieval set as external priors can effectively constrain and guide the generation process, leading to more stable and fine-grained alignment between linguistic semantics and facial geometry.

In this paper, we address the challenge that existing text-driven 3D facial mesh generation and editing methods often fail to accurately capture fine-grained local facial deformations described in long-form text, especially when the descriptions contain multiple localized and direction-sensitive cues. To support fine-grained generation and editing, we first construct FaME-G2E, a multimodal dataset built using a visual LLM and a monocular 3D reconstruction model. FaME-G2E contains fine-grained text--mesh pairs and text--blendshape pairs, providing unified data support for both 3D facial generation and editing tasks. Based on this dataset, we propose RAGMesh, a retrieval-augmented framework that incorporates external facial geometry priors into the generation process. Specifically, we design a Multi-Scale Retrieval Fusion (MSRF) module to retrieve complementary facial priors from both global and regional perspectives and fuse them into a reference mesh within a shared blendshape space. The fused reference mesh provides structured geometric guidance, enabling the model to better infer subtle local deformations that are difficult to learn from text alone. Beyond using retrieval results as additional inputs, RAGMesh further exploits them to guide optimization. We introduce Adaptive RAG-Guided Supervision (AdaRAGS), which identifies text-relevant facial regions according to the retrieved local priors and applies region-aware constraints during training. This design encourages the model to focus on subtle local changes, such as asymmetric mouth movements or eye-region deformations, rather than being dominated by large static facial areas. As a geometry-centric framework, RAGMesh can be seamlessly combined with existing texture synthesis and rendering pipelines. Extensive experiments demonstrate its effectiveness in fine-grained 3D facial mesh generation and editing.

\begin{itemize}
    \item We construct FaME-G2E, a large-scale multimodal dataset containing fine-grained long-form text–mesh annotations and paired text–blendshape samples, which enables supervised learning for both 3D face generation and editing and alleviates the reliance on SDS-based optimization.
    \item We present RAGMesh, a retrieval-augmented framework for controllable 3D facial mesh generation and editing from long-form textual descriptions. By introducing topology-consistent facial geometry priors, RAGMesh bridges complex language descriptions and localized facial deformations, enabling more precise text-driven control over fine-grained facial geometry.
    \item We design a geometry-aware retrieval and supervision mechanism for local facial editing. The proposed MSRF module retrieves and fuses global and region-level facial priors into a reference mesh, while AdaRAGS converts retrieved local deformation cues into region-aware supervision, improving localized editing accuracy and reducing unintended changes in non-target facial regions.
\end{itemize}

\section{Related Works}
\label{sec:2_relatedwork}
\subsection{Text-driven Facial Mesh Generation}
Text-driven facial mesh generation aims to synthesize 3D meshes directly from natural language descriptions. A straightforward approach is to learn a direct mapping from text to a 3D mesh using neural networks~\cite{Codetalke,FaceDiffuser}. However, when processing long and structurally complex textual inputs, the model tends to focus on dominant tokens while neglecting fine-grained semantic details. Due to the scarcity of 3D data, many studies leverage 2D priors for optimization~\cite{Text2Mesh,TADA}. For example, T2P~\cite{T2P} employs a CLIP-based semantic alignment mechanism to jointly optimize facial parameters and textual features, enabling text-driven mesh generation. Score Distillation Sampling (SDS)~\cite{DreamFusion} optimizes 3D representations by acquiring gradient signals from a pretrained text-to-image diffusion model. Methods such as DreamFace~\cite{DreamFace} and FaceG2e~\cite{FaceG2E} adopt the SDS framework to achieve text-driven 3D face generation. Although SDS leverages 2D image gradients for supervision and eliminates the need for large-scale 3D datasets, it primarily captures global semantics and is computationally expensive. To balance efficiency and reliability, ICE~\cite{ICE} employs an LLM-based instruction parser and a semantic-guided low-dimensional solver for predefined blendshape bases. However, its generated results still exhibit limited global coherence.

\subsection{Datasets for Text-driven Mesh Generation}
Text-driven 3D tasks encompass both generation and editing, yet existing datasets are insufficient to support such tasks effectively. Some 3D facial scan datasets, such as BU-3DFE~\cite{YinWSWR06} and BP4D~\cite{ZhangYCCRHLG14}, provide high-quality geometry but lack detailed textual descriptions. Although Describe3D~\cite{DESCRIBE3D} provides fine-grained textual annotations for 3D meshes, its predefined description categories, limited sample size, and task-specific design make it unsuitable for editing. Due to the scarcity of large-scale 3D datasets, most existing methods rely on 2D priors within SDS-based frameworks~\cite{SDS1,DreamFusion,FaceG2E}. However, these SDS-based approaches can handle only short text inputs, struggle with long, structurally complex textual descriptions, and are nearly incapable of performing continuous text-driven editing. By integrating Vision–Language Large Models~\cite{vlm,vlm2} with monocular reconstruction techniques~\cite{EMOCA}, we generate high-quality paired Text-Mesh pair data from large-scale 2D facial images, substantially enriching both data diversity and scale. Moreover, the incorporation of diverse blendshape representations enables FaME-G2E to seamlessly handle both generation and editing tasks.

\subsection{RAG for 3D Content Generation}
Retrieval-Augmented Generation (RAG) has been widely adopted across domains such as natural language understanding, knowledge-grounded text generation, and image synthesis~\cite{SKR,HuangKZ21,HuangKZ211}. It enhances generative models by integrating external knowledge retrieved from large-scale databases. Recently, RAG has been extended to 3D content generation, demonstrating that incorporating external priors can improve both geometric consistency and overall generation quality~\cite{ReMoDiffuse,AMD,RetDream}. ReMoDiffuse~\cite{ReMoDiffuse} dynamically integrates retrieved motion samples with textual semantics for text-to-motion generation, while AMD~\cite{AMD} decomposes textual descriptions and fuses language and motion priors within a diffusion framework. However, these methods primarily operate on temporal motion priors and global semantic structures, without explicitly modeling conflicts among fine-grained local geometric cues under long-form descriptions. Similarly, RetDream~\cite{RetDream} retrieves semantically relevant 3D assets as geometric priors for SDS-based text-to-3D optimization, where the retrieved assets mainly provide global-level geometric guidance. In contrast, facial geometry generation introduces a more challenging setting, where long textual descriptions often involve region-specific and direction-aware constraints that require fine-grained geometric consistency. To address this, we move beyond single-reference retrieval or naive feature aggregation and construct a more expressive geometric prior by jointly leveraging multiple retrieved facial meshes. We perform blendshape-level fusion within a shared deformation space, yielding a structured and descriptive reference mesh. This fusion strategy preserves deformation patterns consistently supported across semantically relevant retrieved samples, while suppressing incompatible local deformations across different facial regions. As a result, the fused representation avoids over-smoothing artifacts commonly observed in vertex-space averaging and provides a more stable and structured geometric prior. Furthermore, the aggregated reference mesh naturally enables region-aware supervision, allowing fine-grained alignment between textual semantics and localized facial deformations, thereby significantly improving both 3D facial mesh generation and editing.

\section{Methodology}

\subsection{FaME-G2E}
\subsubsection{Motivation}
Most existing text-driven 3D generation methods rely on 2D prior
supervision, such as CLIP-based semantic alignment or SDS-based
optimization~\cite{FaceG2E,SDS1}, to circumvent the lack of
large-scale text-annotated 3D datasets. However, these 2D priors
mainly emphasize image-level semantic consistency and provide only
indirect supervision for 3D geometry. In particular, CLIP-based
representations have limited capability in preserving multiple
fine-grained and direction-sensitive constraints contained in long,
structurally complex descriptions. Similarly, SDS derives optimization
signals from pretrained 2D diffusion models through rendered images,
rather than directly constraining localized 3D facial geometry.
Consequently, these approaches often struggle to accurately model
subtle and asymmetric facial deformations, such as unilateral mouth
movements, eyebrow variations, and local cheek contractions. These
limitations hinder the establishment of precise correspondence between
long-form textual semantics and fine-grained facial geometry.

To further clarify the limitations of existing data resources and
supervision paradigms, Table~\ref{data_supervision_comparison}
provides two complementary comparisons. 
As summarized in Table~\ref{data_supervision_comparison},
existing 3D facial datasets primarily provide geometric scans,
blendshape bases, or predefined expression labels, but generally lack
paired free-form textual descriptions. Describe3D introduces text--mesh
annotations, yet its predefined categories, limited scale, and absence
of editing-oriented samples restrict its applicability to long-form
generation and localized editing. Meanwhile, CLIP- and SDS-based
paradigms avoid paired 3D annotations by exploiting pretrained 2D
priors, but their image-level supervision provides only indirect
constraints on localized facial geometry. These limitations motivate
FaME-G2E, which provides large-scale text--mesh pairs for generation
and text--blendshape pairs for fine-grained editing.
\begin{table*}[t]
\centering
\caption{Comparison of existing 3D facial datasets and supervision
paradigms for text-driven generation and editing.}
\label{data_supervision_comparison}

\renewcommand{\arraystretch}{1.08}
\setlength{\tabcolsep}{10pt}
\small

\begin{tabular}{cccc}
\toprule
\multicolumn{4}{c}{\textbf{Dataset Comparison}} \\
\midrule
Dataset
& Geometry Representation
& Text Description
& Annotation Style \\
\midrule

BU-3DFE~\cite{YinWSWR06}
& 3D Mesh
& \xmark
& Emotion \\

FaceWarehouse~\cite{CaoWZTZ14}
& Blendshape
& \xmark
& Expression Parameters \\

Describe3D~\cite{DESCRIBE3D}
& 3D Mesh
& Limited
& Template-based Text \\

\textbf{FaME-G2E}
& 3D Mesh and Blendshape
& \cmark
& Free-form Text \\

\midrule
\multicolumn{4}{c}{\textbf{Supervision Paradigm Comparison}} \\
\midrule

Paradigm
& Supervision
& Optimization Granularity
& Local Control \\
\midrule

CLIP-based
& Image--Text Alignment
& Global-oriented Semantic Alignment
& \xmark \\

SDS-based
& Diffusion Prior
& Global-oriented Semantic Optimization
& Limited \\

LLM-based Parsing
& Semantic Mapping
& Region-level Parameter Mapping
& Limited \\

\textbf{Paired Text--Mesh}
& Text--Mesh Pairs
& Fine-grained Geometric Optimization
& \cmark \\

\bottomrule
\end{tabular}

\vspace{2pt}
\begin{minipage}{0.95\textwidth}
\centering
\footnotesize
\textit{\cmark: supported; \xmark: unsupported; Limited: partially supported.}
\end{minipage}
\end{table*}

\begin{figure*}[t]
 \centering
 \includegraphics[width=2\columnwidth]{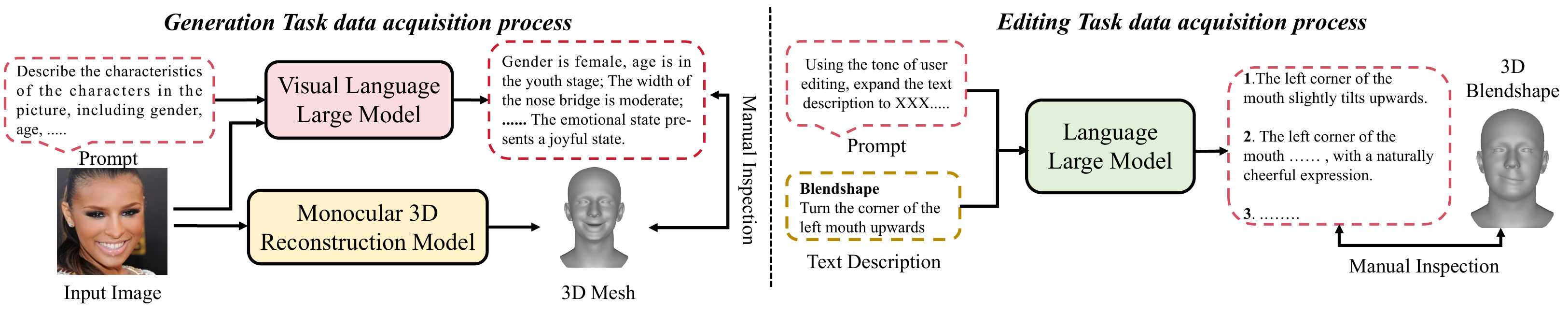}
 \vspace{-3mm}
 \caption{Pipeline of FaME-G2E dataset construction. The left presents the generation task, and the right depicts the editing task.}
 \vspace{-3mm}
 \label{dataset}
\end{figure*}
\subsubsection{Dataset Construction}
As shown in Figure~\ref{dataset}, FaME-G2E comprises two complementary subsets. For the generation task, we use images from AffectNet as visual inputs and employ the vision-language model, doubao-1-5-thinking-vision-pro, to generate fine-grained facial descriptions using predefined visual question-answering prompts. In parallel, we apply EMOCA~\cite{EMOCA} to reconstruct the corresponding 3D facial meshes in the FLAME topology with 5023 vertices, producing paired Text--Mesh samples. After manual validation, only samples with high semantic consistency between the generated text and reconstructed mesh are retained. 
For the editing task, we further construct a Text--Blendshape subset from shape and motion blendshape control bases. For each blendshape basis, we generate and expand user-oriented textual descriptions using LLM-augmented prompts that cover diverse local deformation types, directional attributes, and expression intensities. To simulate different editing intensities, we assign different activation values to the corresponding blend shape controls to represent weak, natural, and strong expressions, respectively. The generated descriptions are manually verified to ensure consistency with the target blendshape semantics. The detailed prompts for both data generation and data editing are provided in the supplementary material (SM).

\begin{table}[!t]
\footnotesize
\centering
\renewcommand{\arraystretch}{1.1}
\setlength{\tabcolsep}{2.8pt}
\caption{Data characteristics in FaME-G2E for different tasks.}
\label{tab:dataset_comparison}
\begin{tabular}{p{2.5cm}p{2.5cm}p{2.5cm}}
\toprule
\textbf{Attribute} & \textbf{Generation Task} & \textbf{Editing task} \\
\toprule
\textbf{Data Scale} & 
500K & 
30K \\
\textbf{Text Length} & 
Avg. 63 words & 
Avg. 11 words \\
\textbf{Annotation Type} & 
\makecell[l]{Emotion, AU, Motion,\\Shape, Attribute} & 
\makecell[l]{Emotion, AU, Motion,\\Shape} \\
\textbf{Intensity} & No intensity & Weak/Natural/Strong \\
\textbf{Annotation Strategy} &Type Mixing  &Type Independent \\
\bottomrule
\end{tabular}
\label{statistics}
\end{table}
\subsubsection{Dataset Statistics}
In summary, FaME-G2E is a large-scale and comprehensive multimodal dataset designed for text-driven 3D facial generation and editing. As shown in Table~\ref{statistics}, it contains 500K Text–Mesh pairs with fine-grained multi-dimensional annotations for generation, and 30K Text-Blendshape pairs curated for editing. The Text-Blendshape subset includes 50 FLAME shape blendshapes, 51 ARKit motion blendshapes, 14 AU-based blendshapes, and 7 standard emotion blendshapes. Each blendshape is further divided into three intensity levels and is associated with 50 to 100 distinct textual descriptions. This subset provides diverse textual descriptions and corresponding blendshape control bases, enabling interpretable and controllable expression modeling and manipulation. Compared with existing 3D facial datasets, FaME-G2E is the first large-scale multimodal paired Text–Mesh dataset that simultaneously supports text-driven generation and editing tasks.

\subsection{RAGMesh}
\begin{figure*}[t]
 \centering
 \includegraphics[width=2\columnwidth]{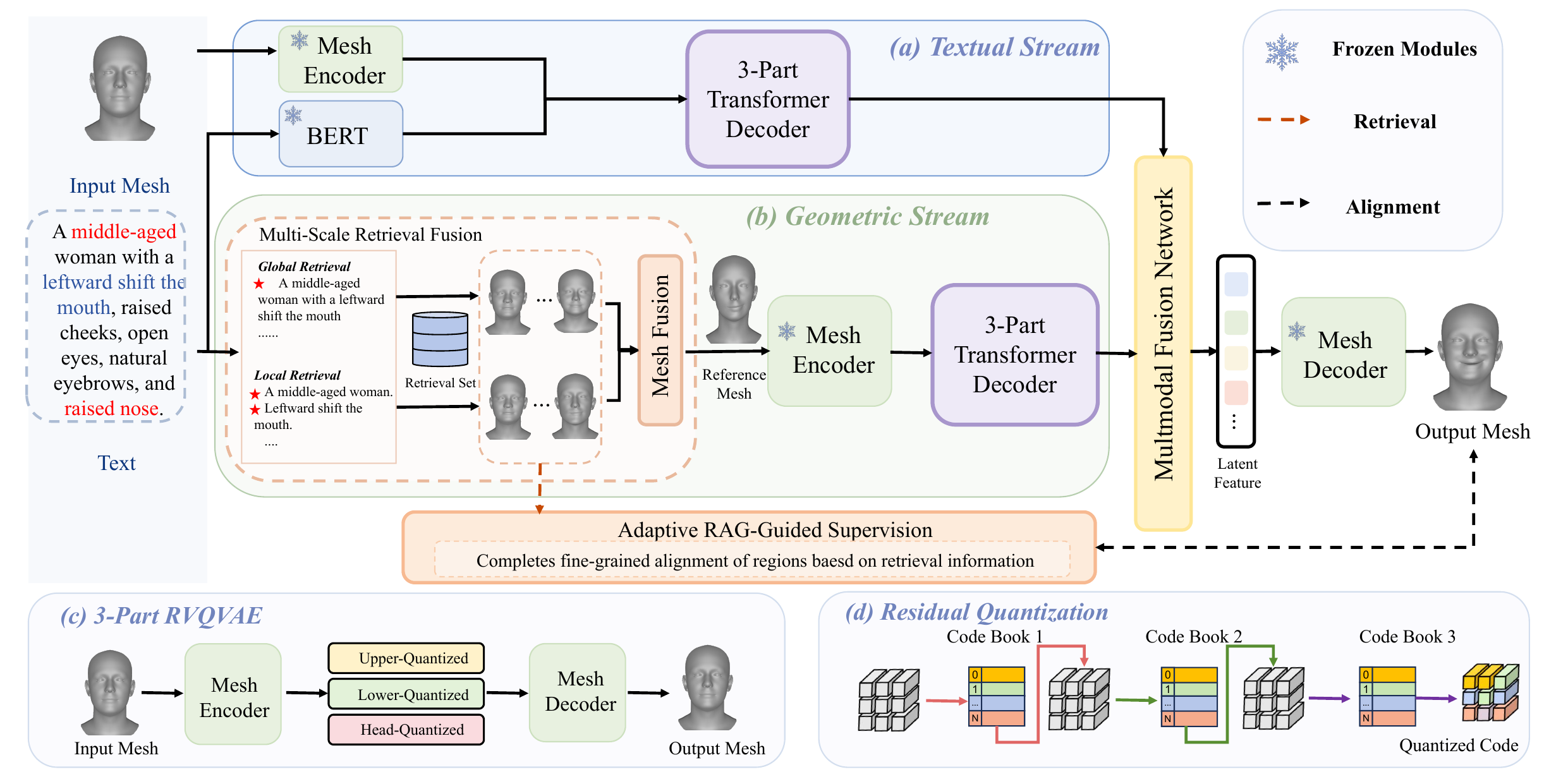}
 \caption{Overview of RAGMesh. (a) (b) show the textual stream and geometric stream in the dual-stream architecture. Given a text description and an input mesh, MSRF retrieves meshes from a predefined retrieval set via global and local semantic retrieval. The retrieved meshes are fused to generate a reference mesh serving as a geometric prior. Next, two 3-Part Transformer decoders process the input mesh with textual and geometric features. Finally, the multimodal fusion network fuses textual and geometric features to generate latent features, which are decoded into the target mesh. AdaRAGS completes fine-grained alignment of regions based on retrieval information. (c) (d) show the architecture of 3-Part RVQVAE and Residual Quantization.}
 \label{pipeline}
 \vspace{-5mm}
\end{figure*}

As illustrated in Figure~\ref{pipeline}, our framework addresses both generation and editing tasks through text-driven 3D facial synthesis. The generation task takes a template as input to produce text-guided meshes, whereas the editing task deforms existing 3D meshes according to the text. The dual-stream architecture leverages the reference mesh generated by Multi-Scale Retrieval Fusion (MSRF) to compensate for the missing local details in long textual descriptions, thereby enhancing the fine-grained geometric representation and semantic consistency. In addition, Adaptive RAG-Guided Supervision (AdaRAGS) explicitly aligns textual semantics with localized geometric structures, improving consistency and robustness.

\subsection{3-Part RVQVAE for Representation}
Build a compact, structured latent representation using a discrete codebook, enabling the model to generate in a controllable, continuous space. Traditional compression models~\cite{GuoMJW024,Codetalke} often fail to preserve subtle local deformations. To address the issue, as shown in Figure~\ref{pipeline} (c), we construct a 3-Part RVQVAE network to encode facial priors and divide the facial mesh into upper face, lower face, and head regions for region-specific compression and reconstruction. Our 3-Part RVQVAE enables the model to focus on local geometric changes and subtle facial deformation modeling. The encoding process can be described as:
\begin{equation}
\setlength\abovedisplayskip{2pt}
\setlength\belowdisplayskip{1pt}
\mathbf{z}_{r}^{(0)} = \mathcal{E}_r(\mathbf{V}_r),
\end{equation}
\begin{equation}
\setlength\abovedisplayskip{1pt}
\setlength\belowdisplayskip{2pt}
\mathbf{z}_{q,r}^{(l)} = \operatorname{Quantization}(\mathbf{z}_{r}^{(l-1)} ),
\end{equation}
where $r \in \{ \text{upper}, \text{lower}, \text{head} \}$, $\mathbf{V}_r$ denotes the vertices of the facial region $r$, 
$\mathcal{E}_r(\cdot)$ is the mesh encoder of the 3-part RVQVAE for region $r$,  and $\mathbf{z}_{q,r}^{(l)}$ denotes the quantized representation from the $l$-th quantized layer. 
As shown in Figure~\ref{pipeline} (d), the residual update allows each quantization layer to encode the remaining error of the previous stage, progressively refining the latent representation:
\begin{equation}
\setlength\abovedisplayskip{2pt}
\setlength\belowdisplayskip{2pt}
\mathbf{z}_{r}^{(l)}=\mathbf{z}_{r}^{(l-1)} - \mathbf{z}_{q,r}^{(l)},
\end{equation}
where $\mathbf{z}_{r}^{(l)}$ is the residual feature at $l$-th quantization layer.

The decoder takes the quantized features of different layers as input and predicts the target facial vertices: 
\begin{equation}
\setlength\abovedisplayskip{2pt}
\setlength\belowdisplayskip{2pt}
\hat{\mathbf{V}}_r = \mathcal{D}_r(\sum_{l=1}^{L}\mathbf{z}_{q,r}^{(l)} ),
\end{equation}
where $\mathcal{D}_r(\cdot)$ denotes the decoder for facial region $r$, $\hat{\mathbf{V}}_r$ represents the predicted vertices of that region. 

\subsection{Multi-Scale Retrieval Fusion}
MSRF aims to generate a reference mesh that semantically aligns with the description. It consists of two stages: multi-scale retrieval and mesh fusion. In the retrieval stage, global and local retrieval are performed simultaneously to capture multi-scale semantic and geometric correlations within a pre-built retrieval set. The global retrieval captures overall facial structure and emotional consistency, while the local retrieval focuses on regional geometry and subtle facial deformations. In the fusion stage, the retrieved meshes are aggregated using a weighted strategy to produce the reference mesh.

To ensure retrieval efficiency and semantic representativeness, we construct a retrieval set $\mathcal{S}$ from FaME-G2E, which includes paired text and mesh. We extract hybrid representations of geometric features and textual semantics, and apply hierarchical clustering to select the most representative samples for the retrieval candidates. Given a text description $T$, a Text2Vec~\cite{HuangPZLLLWSY24} encoder $\mathcal{E}_t$ is leveraged to obtain the text embedding, which serves as a query vector to retrieve the Top-$K$ facial meshes from the retrieval set $\mathcal{S}$. 

We perform both global and local retrieval. Global retrieval operates on the entire text sequence to capture overall semantic alignment. In contrast, local retrieval divides $T$ into $N$ smaller segments based on punctuation marks and performs stepwise retrieval to capture fine-grained, region-specific semantic cues. The retrieval process can be formulated as:
\begin{equation}
\setlength\abovedisplayskip{3pt}
\setlength\belowdisplayskip{1pt}
Q_{g}, \mathcal{R}_g =
\operatorname{TopK}
(\operatorname{Sim}(\mathcal{E}_t(T), \mathcal{E}_t(T_{k})), V_k), k\in{|\mathcal{S}|},
\end{equation}
\begin{equation}
\setlength\abovedisplayskip{1pt}
\setlength\belowdisplayskip{21pt}
Q_{l}, \mathcal{R}_l = 
\bigcup_{n=1}^{N}
\operatorname{TopK}(
\operatorname{Sim}(\mathcal{E}_t(T^{n}),\, \mathcal{E}_t({T_{k}})),V_k),
\end{equation}
where $(T_{k}, V_{k})\!\in\!\mathcal{S}$ is the $k$-th text and mesh pair in the retrieval set,Sim$(\cdot)$ is the similarity operation, and $T^n$ is the $n$-th text segment. $\mathcal{R}_g$ and $\mathcal{R}_l$ are the global and local retrieved meshes. $Q_{g}$ and $Q_{l}$ are the semantic similarity scores with the corresponding text query global and local retrieved mesh.
The fusion weight of the $k$-th retrieved mesh is computed as:
\begin{equation} 
{w}_{b}^{k} = \frac{ \exp \left( \left(q_{b}^{k}+\eta_{b}^{k}\right) / \tau \right) }{ \sum_{j=1}^{K} \exp \left( \left(q_{b}^{j}+\eta_{b}^{j}\right) / \tau \right) }, \quad b \in \{g,l\}, 
\end{equation} where $q_b^k$ is the similarity of the k-th geometric text. $\eta_{b}^{k}\sim\mathcal{N}(0,\sigma^{2})$ is a small stochastic perturbation, and $\tau$ is a temperature parameter controlling the sharpness of the weight distribution.

Finally, the reference mesh is obtained by merging the global and local meshes in the blendshape space:
\begin{equation} \mathbf{V}_{\mathrm{ref}}^{r} = \alpha \mathbf{w}_{g} \mathcal{R}_g + \beta \mathbf{w}_{l} \mathcal{R}_l, 
\end{equation} 
where $\alpha$ and $\beta$ control the contributions of the global and local retrieval branches.

\subsection{Textual Geometric Dual-Stream Architecture}
To better leverage text and input mesh information, we design a dual-stream architecture that jointly processes textual semantics and geometric priors for 3D facial generation and editing. Specifically, the textual stream extracts semantic representations from the textual description, and the geometric stream integrates multi-scale information from the retrieved reference meshes:
\begin{equation}
\setlength\abovedisplayskip{2pt}
\setlength\belowdisplayskip{1pt}
\mathbf{f}_{t,r} = \Phi_{t,r}\!\left([\mathcal{E}_r\left(\mathbf{V}_{inp,r}\right); \mathrm{BERT}(T)]\right),
\end{equation}
\begin{equation}
\setlength\abovedisplayskip{2pt}
\setlength\belowdisplayskip{2pt}
\mathbf{f}_{g,r} = \Phi_{g,r}\!\left([\mathcal{E}_r\left(\mathbf{V}_{inp,r}\right); \mathcal{E}_r\left(\mathbf{V}_{ref,r}\right)]\right),
\end{equation}
where $\mathbf{V}_{inp,r}$ and $\mathbf{V}_{ref,r}$ denotes vertices of facial region $r$ for the input mesh and reference mesh, respectively. $\Phi_{t,r}$ and $\Phi_{g,r}$ represent 3-part Transformer decoder for processing textual and geometric references, and $\mathbf{f}_{t,r}$ and $\mathbf{f}_{g,r}$ denote the textual and geometric encoding features of the facial region. $\mathbf{f}_{t,r}$ and $\mathbf{f}_{g,r}$ are integrated via a Multimodal Fusion Network (MFN), a three-layer MLP with GELU activations designed to fuse the concatenated text-guided and geometry-guided latent features, to produce a fused latent representation that is subsequently decoded into the target mesh:
\begin{equation}
\setlength\abovedisplayskip{2pt}
\setlength\belowdisplayskip{1pt}
\mathbf{z}_{f,r} = \mathrm{MFN}\!\left([\mathbf{f}_{t,r};\mathbf{f}_{g,r}]\right),
\end{equation}
\begin{equation}
\setlength\abovedisplayskip{1pt}
\setlength\belowdisplayskip{2pt}
\hat{\mathbf{V}}_r = \mathcal{D}_r(\mathbf{z}_{f,r} ),
\end{equation}
where $\mathbf{z}_{f,r}$ denotes the fused  representation of region $r$, $\hat{\mathbf{V}}_r$ is the predicted mesh vertex of facial region $r$.

\subsection{Adaptive RAG-Guided Supervision}
Text descriptions usually convey localized and directional facial semantics. Directly applying global supervision fails to capture fine-grained regional cues and deformation directions. To address this issue, we propose AdaRAGS to enhance supervision on the vertices of specific regions.
It leverages fine-grained, semantically relevant meshes retrieved by the MSRF, enabling directionally consistent, semantically aligned local supervision. Firstly, we define the facial vertex mask:
\begin{equation}
\setlength\abovedisplayskip{2pt}
\setlength\belowdisplayskip{2pt}
\mathcal{M}
= \bigcup_{m=1}^{M} 
\Big\{\, 
i \;\big|\;
\big\|
\mathcal{R}_{l,i}^{m} - \mathbf{V}_{0,i}
\big\|_2 > \tau,
\,\Big\},
\end{equation}
where $\mathcal{M}$ denotes a mask set that selects vertices with significant local motion, $M$ represents the total number of locally retrieved reference meshes, $i$ denotes the index of a vertex, $\mathcal{R}_l^{m}$ is the $m$-th retrieved mesh, $\mathrm{\mathbf{V}_0}$ is the template mesh, $\tau$ is the displacement threshold. We apply the L1 loss only to the masked vertices to guide the model in learning semantically relevant local deformations:
\begin{equation}
\setlength\abovedisplayskip{2pt}
\setlength\belowdisplayskip{2pt}
\mathcal{L}_{M} = 
\frac{1}{M}
\sum_{m=1}^{M}
\frac{\sum_{i \in \mathcal{M}^{(m)}}
\big\|
{\mathbf{V}}_{i} - \hat{\mathbf{V}}_{i}
\big\|_1}{|\mathcal{M}^{(m)}|},
\label{eq:mask_l1_loss}
\end{equation}
where $\mathcal{M}^{(m)}$ denotes the vertex mask corresponding to the $m$-th locally retrieved mesh, $\mathbf{V}_{i}$ and $\hat{\mathbf{V}}_{i}$ denote the i-th vertex of ground truth and predicted vertex, encouraging the network to focus on local motion patterns.

\subsection{Loss functions}
We first train the 3-Part RVQVAE to reconstruct different regions of the facial mesh. Each model is optimized using a facial mesh reconstruction loss, along with two intermediate latent-level losses.
\begin{equation}
\begin{aligned}
\mathcal{L}_{\mathrm{rvq}}
&= \sum_{r}\|\mathbf{V}_r - \hat{\mathbf{V}}_r\|_1 + \sum_{r,l}\Big(
\|\operatorname{SG}(\hat{\mathbf{z}}_{r}^{\,l-1}) - \mathbf{z}_{q,r}^{\,l}\|_2^2
\\
&+ \delta\,\|\hat{\mathbf{z}}_{r}^{\,l-1}
- \operatorname{SG}(\mathbf{z}_{q,r}^{\,l})\|_2^2
\Big),
\end{aligned}
\end{equation}
where the first term is a mesh reconstruction loss, $\rm{SG}$ stands for a stop-gradient operation, and $\delta$ refers to the weighting factor controlling the update rate. 

For both generation and editing tasks, we train RAGMesh using the corresponding data from FaME-G2E. We optimize our model through facial mesh reconstruction loss, intermediate latent space loss, and AdaRAGS:
\begin{equation}
\setlength\abovedisplayskip{2pt}
\setlength\belowdisplayskip{2pt}
\mathcal{L}
 = \sum_{r}\|\mathbf{V}_r - \hat{\mathbf{V}_r}\|_2^2
+ \sum_{r,l}\|\hat{\mathbf{z}}_{r}^{\,l-1}
- \operatorname{SG}(\mathbf{z}_{q,r}^{\,l})\|_2^2
+ \gamma\ \mathcal{L}_{M},
\end{equation}
where $\gamma$ is a weight parameter for the AdaRAGS.

\section{Experments}
\subsection{Implementation details}
Our framework is built on the PyTorch platform and trained on RTX 6000 Ada. All experiments use a random 8:1:1 train, val, and test split on the proposed dataset. First, we optimize the 3-part RVQVAE, with each RVQVAE setting the layer numbers to 8 and latent spatial dimensions to 1024. The hyperparameter $\delta$ equals 0.1. Adam optimizer is used to train RVQVAE with a learning rate of 1$\times$ $10^{-5}$ and a batch size of 8 for 200 epochs. Subsequently, we optimize the dual-stream architecture by setting the hidden dimension to 1024. Both the number of heads and the number of layers are set to 8. In the retrieval fusion module, the retrieval size is set to TopK=5, and the weighting factors for the reference meshes are $\alpha$=0.3 and $\beta$=0.7. In the Adaptive RAG-Guided Supervision, the threshold parameter is set to $\tau$=0.0001. The model is trained for 20 epochs with a batch size of 8, a learning rate of 1 $\times$ $10^{-5}$, and a loss weight $\gamma$=4.

\subsection{Quantitative Evaluation}
To verify RAGMesh, we evaluate all models on both generation and editing tasks using LVE (Loss of Vertex Error), MRE (Mask-based Reconstruction Error), VS (Vendi Score), and TIAS (Text–Image Alignment Score) metrics. 

\subsubsection{LVE} Since RAGMesh performs long-context, fine-grained text-to-3D facial generation, where the input description explicitly specifies local facial regions, deformation directions, and deformation strength, the generation process becomes strongly constrained and exhibits reduced semantic ambiguity. Accordingly, we apply Loss of Vertex Error(LVE) to evaluate the geometric fidelity of generated meshes with respect to text-specified facial deformations:
\begin{equation}
\setlength\abovedisplayskip{1pt}
\setlength\belowdisplayskip{1pt}
\mathrm{LVE} = ||\mathbf{V} - \hat{\mathbf{V}}||_2,
\end{equation}
where $\mathbf{V}$ and $\hat{\mathbf{V}}$ denote ground truth and predicted vertex.

\subsubsection{MRE} To evaluate region-specific generation quality under textual instructions, we adopt Mask-based Reconstruction Error (MRE), which focuses on local structural fidelity. 
MRE measures the geometric discrepancy between the generated mesh and ground-truth mesh within the masked facial regions:
\begin{equation}
\setlength\abovedisplayskip{1pt}
\setlength\belowdisplayskip{1pt}
\mathrm{MRE} =  
\frac{1}{M}
\sum_{m=1}^{M}
\frac{\sum_{i \in \mathcal{M}^{(m)}}
\big\|
{\mathbf{V}}_{i} - \hat{\mathbf{V}}_{i}
\big\|_1}{|\mathcal{M}^{(m)}|}.
\label{eq:mask_l1_loss}
\end{equation}

\subsubsection{VS} Vendi score(VS) measures the effective number of distinct samples based on their pairwise similarity structure, providing a distribution-level assessment of diversity. 
Unlike geometry accuracy metrics, VS captures the global variability of the generated mesh distribution and penalizes mode collapse. 
We generate 10 meshes under the same input condition and compute VS over these samples to evaluate geometric diversity. Formally, given a set of generated meshes $\{V_i\}_{i=1}^{10}$, we construct a similarity matrix:
\begin{equation}
K_{ij} = \mathrm{Sim}(V_i, V_j), \quad i,j = 1, \dots, 10,
\end{equation}
which is then normalized as:
\begin{equation}
\tilde{K} = \frac{K}{\mathrm{tr}(K)}.
\end{equation}

Let $\{\lambda_i\}$ denote the eigenvalues of $\tilde{K}$. VS is defined as:
\begin{equation}
\mathrm{VS} =
\exp\left(
-\sum_{i=1}^{10} \lambda_i \log \lambda_i
\right),
\quad \lambda_i \in \mathrm{eig}(\tilde{K}),
\end{equation}
where $\mathrm{Sim}(\cdot)$ denotes a pairwise mesh similarity function, $\tilde{K}$ is the normalized similarity matrix, $\mathrm{tr}(\cdot)$ is the matrix trace, and $\mathrm{eig}(\cdot)$ denotes eigenvalue decomposition.

\begin{figure}[htb]
 \centering
 \includegraphics[width=1\columnwidth]{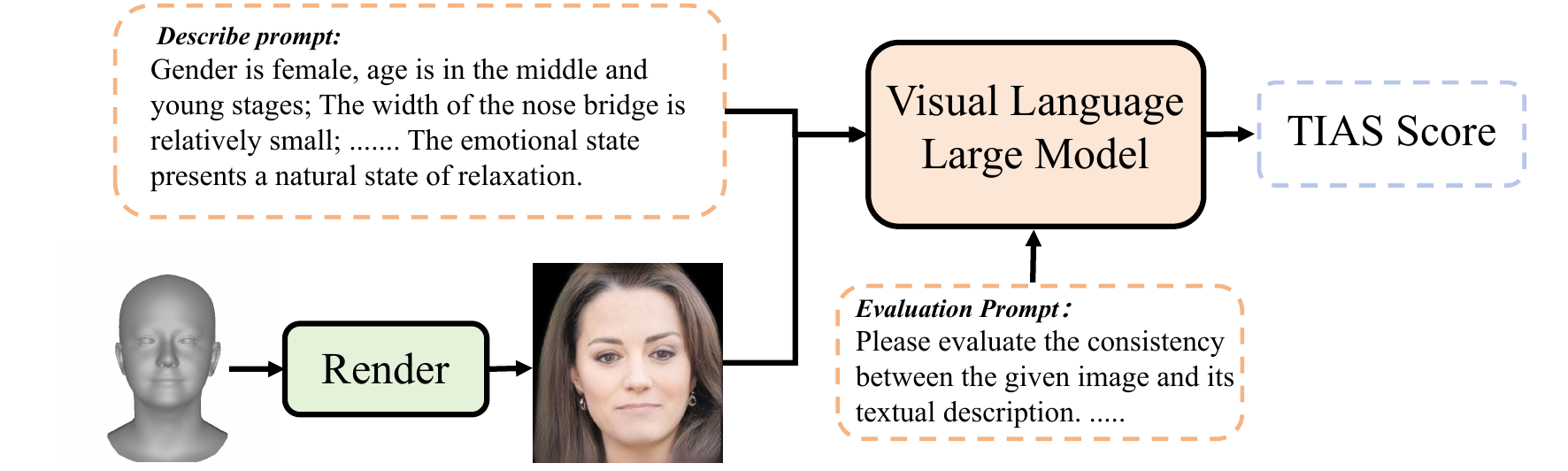}
 \caption{Overview of the TIAS assessment process. Given a generated 3D face, we first render it into a 2D image and then use a visual-language large model to evaluate the consistency between the rendered image and the textual description, producing the final TIAS score.}
 \label{tias}
\end{figure}

\subsubsection{TIAS} Due to the limited capability of CLIP-based similarity models in capturing long-context semantic dependencies, we introduce the Text–Image Alignment Score (TIAS) to evaluate fine-grained text–geometry consistency in long-text scenarios.
Formally, TIAS is computed by a vision-language model-based evaluator. We use condition-driven 3DGS~\cite{GAGAvatar} as our rendering module:
\begin{equation}
\mathrm{TIAS} = f_{\mathrm{VLLM}}\big(\mathbf{I}, T, p_e\big),
\end{equation}
where $\mathbf{I} = \mathrm{Render}(\hat{\mathbf{V}}, \mathbf{I_{s}})$ denotes the rendered image of the generated mesh $\hat{\mathbf{V}}$, $\mathbf{I_{s}}$ is source image, $T$ is the description text, and $p_e$ is the evaluation prompt. Specifically, each generated 3D mesh is rendered into 2D images using 3D Gaussian Splatting (3DGS)\cite{GAGAvatar}, conditioned on the original AffectNet image as the source image. A VLLM-based evaluator is then employed to assess the consistency between the rendered images and the corresponding textual descriptions. Figure~\ref{tias} illustrates the TIAS computation pipeline.

\begin{figure*}[t]
    \centering
    \includegraphics[
        width=0.95\textwidth,
        height=0.75\textheight,
        keepaspectratio
    ]{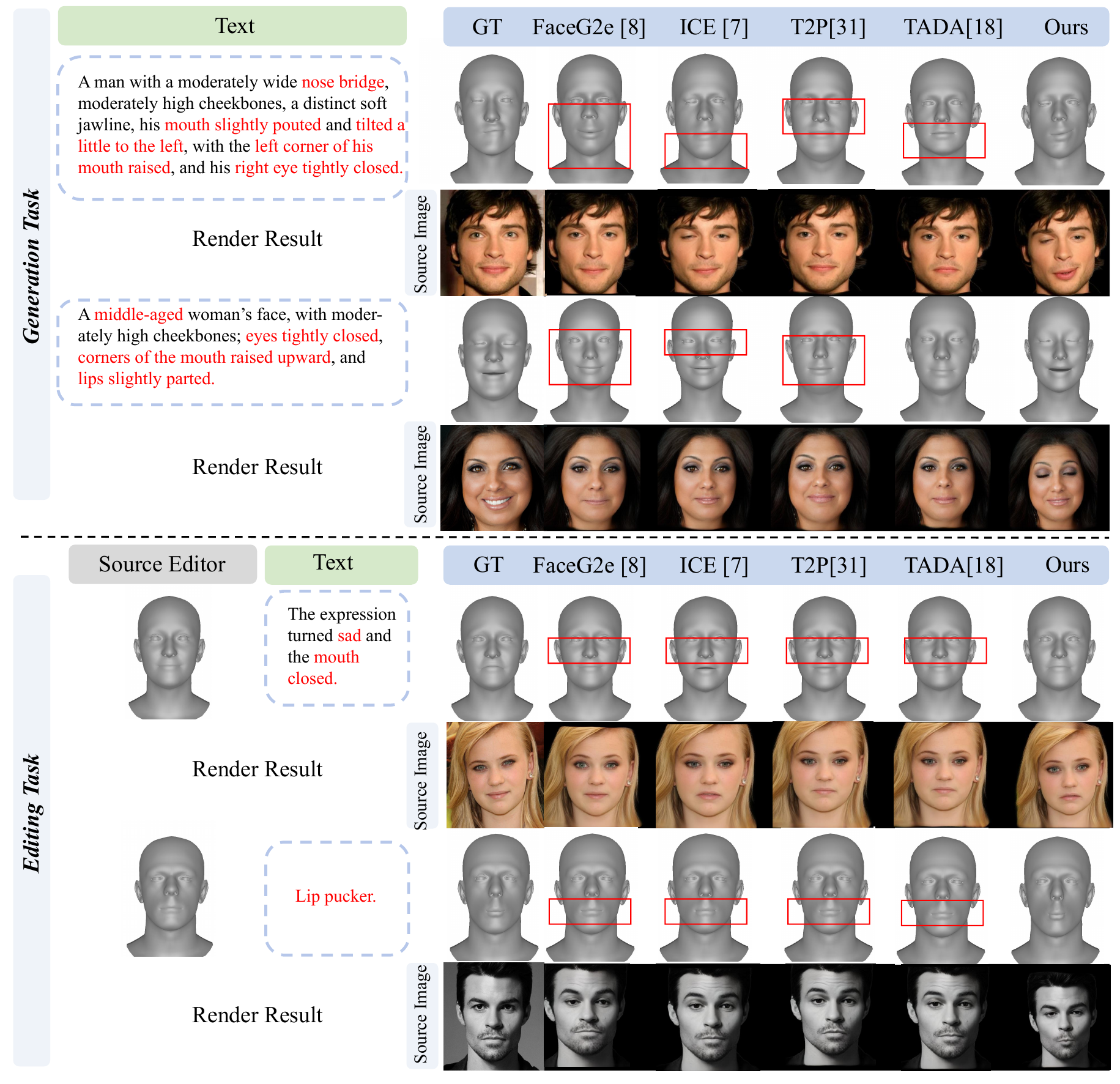}
    \vspace{-3mm}
    \caption{
    Qualitative comparisons of generation and editing results. 
    The upper four rows present generation results and their renderings, while the lower four rows present editing results and their renderings. 
    Failure cases are highlighted with red boxes.
    }
    \vspace{-3mm}
    \label{exp}
\end{figure*}

\subsection{Comparison with State-of-the-Art Methods}
We compare RAGMesh with representative methods for text-driven 3D facial generation and editing. These baselines include FaceG2E~\cite{FaceG2E} and ICE~\cite{ICE}, which are specifically designed for 3D facial modeling, as well as T2P~\cite{T2P} and TADA~\cite{TADA}, two general SDS-based text-to-3D approaches adapted to the facial domain. For T2P and TADA, we retain their original text-guided optimization strategies while reformulating their geometric representations in the FLAME parameter space to enable facial mesh generation and editing. Due to the lack of large-scale 3D facial mesh datasets with long-form textual descriptions, all methods are evaluated on the FaME-G2E dataset using the same data split and experimental settings. This unified protocol facilitates a fair comparison between dedicated facial modeling approaches and adapted general-purpose text-to-3D methods.

Experimental results are reported in Table~\ref{comparison}. As illustrated, RAGMesh consistently outperforms FaceG2E~\cite{FaceG2E}, T2P~\cite{T2P}, TADA~\cite{TADA}, and ICE~\cite{ICE} across all evaluation metrics. Specifically, the improvement in LVE validates enhanced stability in geometric reconstruction, while the lowest MRE demonstrates superior local geometric accuracy. To evaluate generation diversity, we generate ten meshes for each input text and quantify diversity using the Vendi Score. Each generated mesh is rendered into a 2D image via GAGAvatar~\cite{GAGAvatar}, and semantic consistency is assessed using a VLLM-based evaluator. RAGMesh achieves clear improvements in both Vendi Score and TIAS, indicating that the generated facial meshes better preserve fine-grained textual semantics while producing richer and more plausible geometric variations. 

For editing tasks, the diversity is relatively lower than that of generation tasks, since editing instructions usually provide more deterministic constraints on facial regions and deformation directions. Nevertheless, the retrieval-augmented geometric priors introduced by RAGMesh improve the accuracy of text-specified regions, leading to better local fidelity and more controllable editing results. Additionally, inference efficiency is analyzed. FaceG2E~\cite{FaceG2E}, T2P~\cite{T2P}, and TADA~\cite{TADA} require iterative optimization, resulting in slower inference speed. ICE~\cite{ICE} achieves faster inference but still relies on multi-stage parsing and refinement. In contrast, RAGMesh achieves high-quality generation with the highest inference efficiency. These advantages across all metrics collectively demonstrate that RAGMesh outperforms existing methods in fine-grained modeling, semantic consistency, generation diversity, local editing accuracy, and inference efficiency.

Visual comparisons of different methods for facial generation and editing are presented in Figure \ref{exp}. For the generation task, FaceG2e~\cite{FaceG2E}, T2P~\cite{T2P}, and TADA~\cite{TADA} utilize CILP to provide optimized gradients that only capture the global facial structure but fail to model fine-grained details. The LLM-parsing-based ICE~\cite{ICE} method generates meshes with locally accurate textual alignment yet suffers from compromised global coherence. Leveraging the proposed MSRF and AdaRAGS, our RAGMesh effectively focuses on semantically relevant regions via geometric priors and dynamic region-level alignment, enabling more precise modeling of local directional deformations. 

Regarding the editing task, since textual descriptions are typically shorter than in generation, ICE~\cite{ICE} effectively captures local semantic-geometric correspondences, yielding notable reconstruction gains. While FaceG2e~\cite{FaceG2E} improves in this task, it still struggles to capture fine-grained facial deformations. In addition, we integrate the generated meshes with an existing rendering framework to render them into 2D images, and employ a VLLM-based evaluator to assess the consistency between the input text and the generated results. As reported in Table~\ref{comparison}, the results demonstrate that our generated meshes are compatible with downstream rendering pipelines. Meanwhile, the higher text-alignment scores further indicate that RAGMesh achieves more semantically consistent generation and editing results compared with existing methods.

\begin{table*}[htbp]
\centering
\vspace{-2mm}
\caption{Comparison with SOTA Methods. $\downarrow$ means the lower the better and $\uparrow$ means the higher the better.}
\setlength\tabcolsep{4pt}
\resizebox{\textwidth}{!}{%
\begin{tabular}{c c c c c| c c c c| c}
\hline
\multirow{3}{*}{Method}  & \multicolumn{4}{c}{Generation} & \multicolumn{5}{c}{Editing} \\
\cline{2-10}
&LVE\(\downarrow\) &MRE\(\downarrow\) &VS\(\uparrow\) &TIAS\(\uparrow\) &LVE\(\downarrow\) &MRE\(\downarrow\) &VS\(\uparrow\) &TIAS\(\uparrow\)  &Time\(\downarrow\)\\
& ($\times 10^{-6}$) & ($\times 10^{-4}$) & ($\times 10^{-1}$) & ($\times 10^{-1}$) & ($\times 10^{-7}$) & ($\times 10^{-4}$) &($\times 10^{-1}$)  &($\times 10^{-1}$) &s\\ 
\hline 
T2P ~\cite{T2P} &9.46 &13.94 &0.31 &5.51  &4.62 &9.60 &0.18  &6.11 & 176\\
TADA ~\cite{TADA} &8.32 &13.48 &0.38 &5.17  &4.75 &9.54 &0.16  &5.97 &294 \\
FaceG2e~\cite{FaceG2E}   &7.32 &12.71 &0.33  &6.47 &4.15 &9.36 &0.15 &6.15 &361 \\
ICE~\cite{ICE} &10.12 &12.24 &0.36  &7.51 &3.65 &3.64 &0.17  &7.71 &32\\
Ours  &\textbf{1.07} &\textbf{7.88} &\textbf{0.43} &\textbf{8.31} &\textbf{1.72}&\textbf{3.41}&\textbf{0.21} &\textbf{8.59}&\textbf{2}\\
\hline 
\end{tabular}
}
\label{comparison}
\end{table*}

\begin{table*}[t]
\centering
\caption{Ablation experiments regarding model component, retrieval fusion strategy, MFN input setting, and retrieval robustness.}
\vspace{-2mm}
\resizebox{\textwidth}{!}{%
\begin{tabular}{c c c c c| c c c c}
\hline
\multirow{3}{*}{Method}  & \multicolumn{4}{c|}{\textbf{Generation}} & \multicolumn{4}{c}{\textbf{Editing}} \\
\cline{2-9}
&LVE\(\downarrow\) &MRE\(\downarrow\) &VS\(\uparrow\) &TIAS\(\uparrow\) &LVE\(\downarrow\) &MRE\(\downarrow\) &VS\(\uparrow\) &TIAS\(\uparrow\)  \\
& ($\times 10^{-6}$) & ($\times 10^{-4}$) & ($\times 10^{-1}$) & ($\times 10^{-1}$) & ($\times 10^{-7}$) & ($\times 10^{-4}$) &($\times 10^{-1}$)  &($\times 10^{-1}$)\\ 
\hline
\multicolumn{9}{c}{\textbf{Model Component}} \\
\hline
Baseline  &1.35 &8.94 &0.35 &6.76 &1.87 &3.71 &0.15 &6.52  \\
Baseline+MSRF  &1.19&8.89 &\textbf{0.44} &8.15 &1.83 &3.53 &0.21 &8.02 \\
Baseline+MSRF+AdaRAGS  &\textbf{1.07} &\textbf{7.88} &0.43 &\textbf{8.31} &\textbf{1.72}&\textbf{3.41}&\textbf{0.21} &\textbf{8.59} \\
\hline
\multicolumn{9}{c}{\textbf{Retrieval Fusion Strategy}} \\
\hline
Seq-GLF  &1.42 &9.72 &0.40 &7.64  &1.91 &3.73 &0.19 &6.98\\
Single-GF  &1.24 &8.91 &0.41 &7.95  &1.82 &3.69 &0.20 &7.12\\
Single-LF  &1.12 &8.64 &0.41  &8.01  &1.76 &3.52 &0.20 &7.84\\
Single-GLF &\textbf{1.07} &\textbf{7.88} &\textbf{0.43} &\textbf{8.31} &\textbf{1.72}&\textbf{3.41}&\textbf{0.21} &\textbf{8.59}\\
\hline
\multicolumn{9}{c}{\textbf{MFN Input Setting}} \\
\hline
Textual Stream  &1.35 &8.94 &0.39 &6.76 &1.87 &3.71 &0.17 &6.52\\
Geometric Stream  &1.29 &7.92 &0.41 &7.86 &1.77 &3.61 &0.19 &7.91\\
Textual+Geometric Streams &\textbf{1.07} &\textbf{7.88} &\textbf{0.43} &\textbf{8.31} &\textbf{1.72}&\textbf{3.41}&\textbf{0.21} &\textbf{8.59}\\
\hline
\multicolumn{9}{c}{\textbf{Retrieval Robustness}} \\
\hline
OOD Prompt   &1.32 &8.16 &0.40 &6.18 &2.05 &3.77 &0.21 &7.52\\
Random Retrieval &3.51 &9.64  &\textbf{0.65} &4.72 &2.91 &3.94 &\textbf{0.26}  &6.51\\
Ours &\textbf{1.07} &\textbf{7.88} &0.43 &\textbf{8.31} &\textbf{1.72}&\textbf{3.41}&0.21 &\textbf{8.59}\\
\hline
\end{tabular}
}
\label{ablation}
\end{table*}

\begin{figure}[t]
 \centering
 \includegraphics[width=0.9\columnwidth]{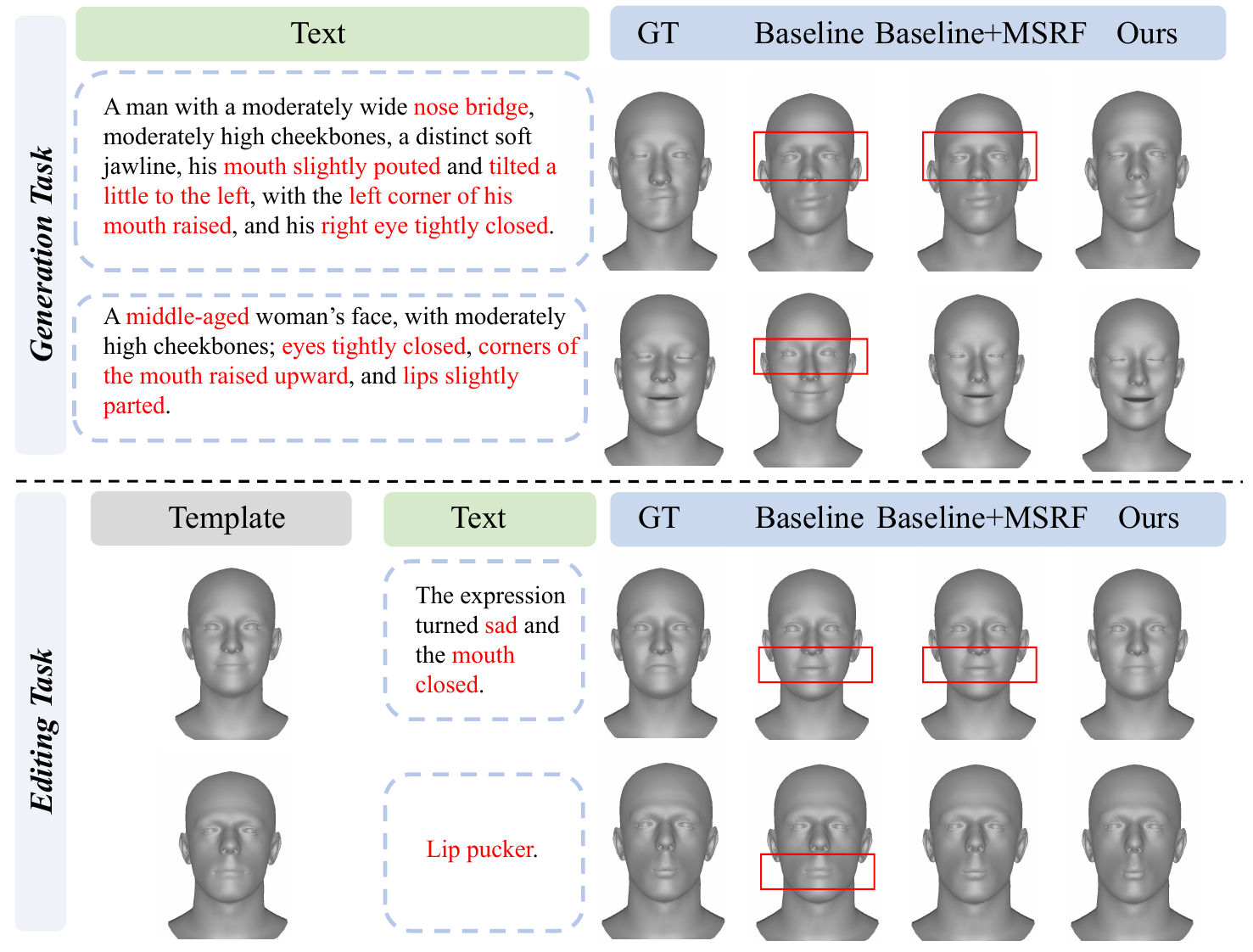}

 \caption{Visualization results of ablation experiments with different modules. Our method better captures fine-grained textual semantics in both global generation and local editing tasks, especially around the eyes, mouth, and expression-related regions highlighted by red boxes.}
 \vspace{-3mm}
 \label{abs1}
\end{figure}

\subsection{Ablation Study for RAGMesh}
To systematically evaluate the contribution of each design choice, we conduct comprehensive ablation studies on model key components, retrieval fusion strategies, MFN input configurations, retrieval robustness, and RVQVAE partitioning settings. We adopt an 8-layer Transformer with 8 attention heads as the baseline model, and leverage BERT to provide the textual representations.

\textbf{Model key Component.} 
MSRF and AdaRAGS are two core components of RAGMesh for reducing ambiguity and preserving local details in long-form text-to-3D facial generation. Notably, our text-only Transformer baseline already outperforms existing state-of-the-art methods, mainly due to the paradigm shift from CLIP-based image-level gradient optimization to direct geometry-level optimization in the 3D mesh/FLAME space. Compared with CLIP-guided supervision obtained from rendered images, direct 3D losses provide more stable and explicit constraints for facial geometry. However, this baseline still struggles to reconstruct subtle facial details and lacks the ability to explicitly distinguish region-specific and direction-aware textual descriptions, as shown in Figure~\ref{abs1}.

To address this limitation, MSRF retrieves complementary global and local geometric cues and fuses them in the blendshape deformation space rather than directly averaging vertices. Unlike single-reference retrieval methods that tend to capture only the dominant mode of facial configuration under long-form ambiguous descriptions, multi-scale retrieval provides a richer and more complete support set for geometry generation. The fusion is performed in a structured deformation space, which preserves consistent geometric components while suppressing conflicting deformations, thereby avoiding the over-smoothing effect of direct geometric averaging. This design is grounded in the observation that text-to-3D facial editing is an inherently ill-posed problem, where multiple geometrically valid solutions may correspond to the same textual instruction. In this setting, the retrieved reference mesh should not be interpreted as a deterministic target, but rather as a data-driven geometric prior that constrains the feasible deformation space. By injecting structural facial cues from the data distribution into the editing process, MSRF reduces ambiguity in the text-to-geometry mapping and improves controllability while preserving multi-modal facial variations.

AdaRAGS further performs mask-guided semantic-geometry alignment, which enforces region-aware consistency between textual semantics and localized deformation fields. Unlike global vertex-level supervision that is dominated by large facial regions, this design explicitly reallocates learning signals to semantically critical regions such as the eyes, eyebrows, and mouth corners. This is not a simple weighting heuristic, but a structured alignment constraint that improves both reconstruction fidelity and editing controllability.

As summarized in Table~\ref{ablation}, incorporating MSRF consistently improves LVE, VS, and TIAS, confirming the effectiveness of retrieval-enhanced geometric priors. Adding AdaRAGS further improves MRE, demonstrating stronger local reconstruction fidelity in text-relevant masked regions. The visual comparisons in Figure~\ref{abs1} further show that, in both generation and editing tasks, the combination of MSRF and AdaRAGS substantially reduces generation ambiguity and better preserves fine-grained facial deformations specified by long textual descriptions.

\begin{figure*}[t]
 \centering
 \includegraphics[width=2\columnwidth]{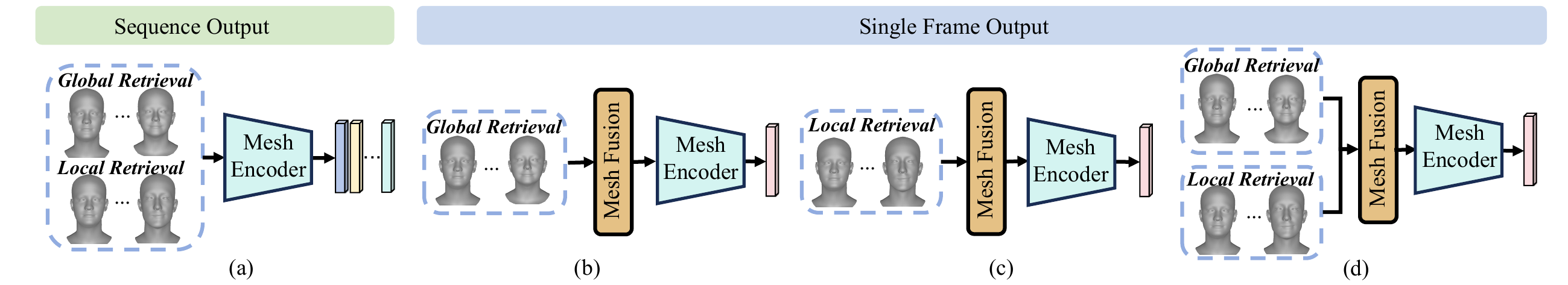}
 \vspace{-3mm}
 \caption{Retrieval fusion. (a) Seq-GLF: Global-Local Fusion. (b) Single-GF: Global Fusion. (c) Single-LF: Local Fusion. (d) Single-GLF: Global-Local Fusion.}
  \vspace{-3mm}
 \label{ablation1}
\end{figure*}

\textbf{Retrieval Fusion Strategy.}
We design four fusion strategies, as shown in Figure~\ref{ablation1}, to investigate how globally and locally retrieved meshes should be integrated. 
Among them, Seq-GLF follows a ReDream~\cite{RetDream} and AMD~\cite{AMD}-style retrieval paradigm, where globally and locally retrieved meshes are provided as sequential reference priors during generation. In contrast, Single-GF, Single-LF, and Single-GLF aggregate the retrieved priors into a single reference mesh using global-only, local-only, and joint global-local fusion, respectively.

As reported in Table~\ref{ablation}, using a single fused reference mesh achieves more stable performance on fine-grained generation and editing tasks, as it avoids the inter-reference inconsistency and noise introduced by sequentially using multiple retrieved meshes. Moreover, Single-GLF outperforms both Single-GF and Single-LF, demonstrating that joint global-local fusion effectively captures both holistic facial structure and text-specified local details. These results validate the necessity of our multi-scale fusion design: rather than directly relying on sequentially retrieved references as in ReDream-style retrieval, RAGMesh resolves conflicts among multi-scale priors and produces a compact, coherent geometric reference for controllable facial generation and editing.

\textbf{MFN Input Setting.} 
Multi-scale retrieved meshes provide geometric priors that complement textual semantics. To examine the effectiveness of the MFN input design, we evaluate different input configurations, including text-only, retrieval-only, and joint text-retrieval inputs. As shown in Table~\ref{ablation}, jointly using textual semantics and retrieval-based geometric priors consistently achieves the best performance across LVE, MRE, VS, and TIAS. 
This demonstrates that textual features provide semantic guidance, while retrieved meshes offer explicit geometric constraints. Their combination enables MFN to produce more accurate and semantically aligned facial generation and editing results.

\begin{figure}[htb]
 \centering
 \includegraphics[width=0.9\columnwidth]{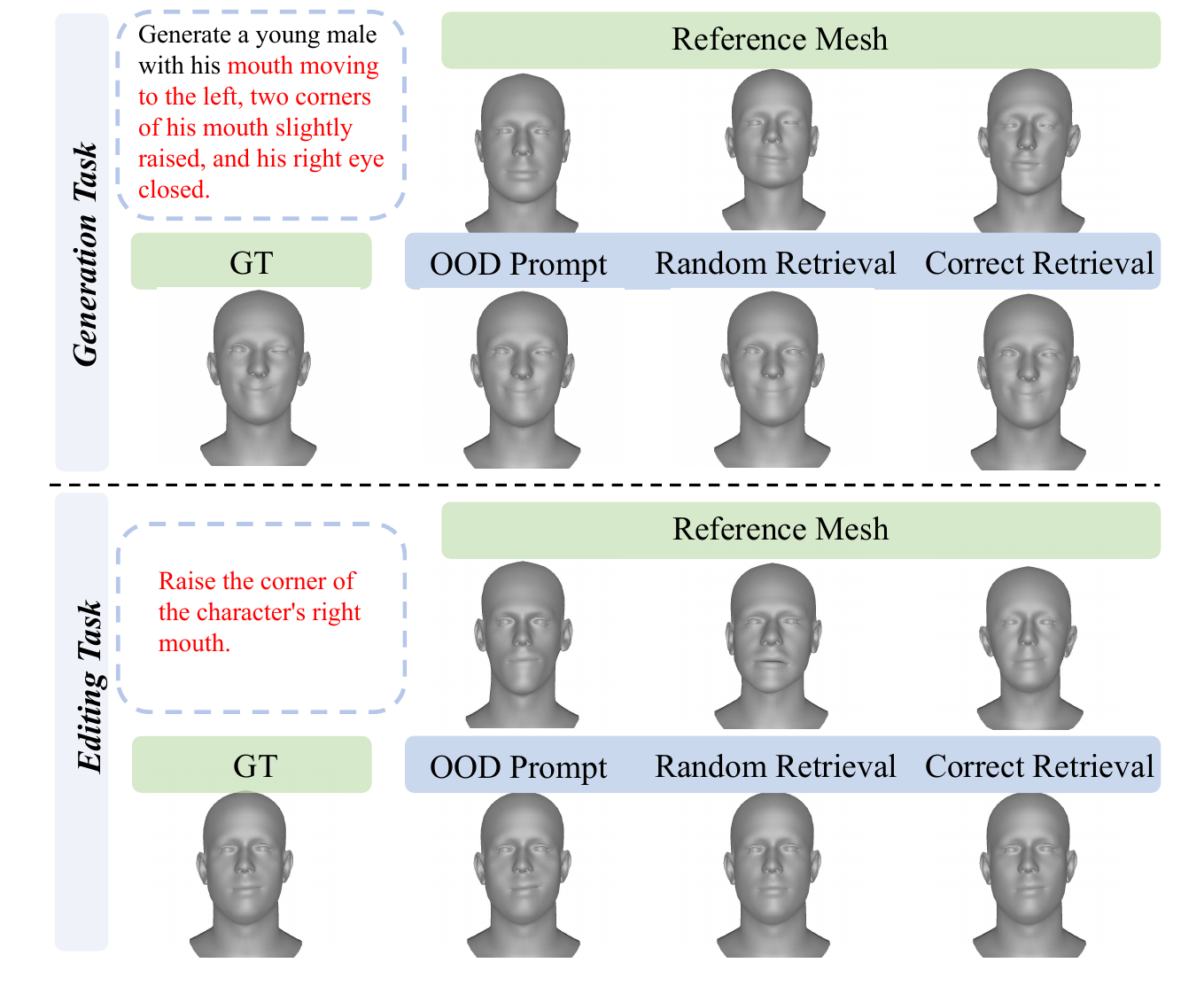}
  \vspace{-3mm}
 \caption{Ablation study on OOD prompts and retrieval quality. }
 \label{abs_rr}

\end{figure}

\textbf{Retrieval Robustness.} 
We further evaluate the robustness of RAGMesh during the retrieval process by using LLM to rewrite the test text as out-of-distribution (OOD) samples and comparing the performance with that of randomly generated samples. As shown in Fig.~\ref{abs_rr}, RAGMesh produces plausible approximations under OOD prompts, indicating its robustness to linguistic variations. When randomly retrieved meshes are used, the editing accuracy decreases, but the generation does not collapse, suggesting that the model is not overly dependent on retrieval quality. In contrast, using correctly retrieved meshes achieves the best performance, demonstrating the importance of accurate retrieval for fine-grained text-guided editing. We further quantitatively evaluate these effects in Table~\ref{ablation}. Results indicate that random retrieval improves output diversity, whereas correct retrieval prioritizes fidelity, highlighting a clear trade-off between diversity and accuracy in retrieval-conditioned generation.

\subsection{User Study Evaluation}
\begin{figure}[t]
 \centering
 \includegraphics[width=1\columnwidth]{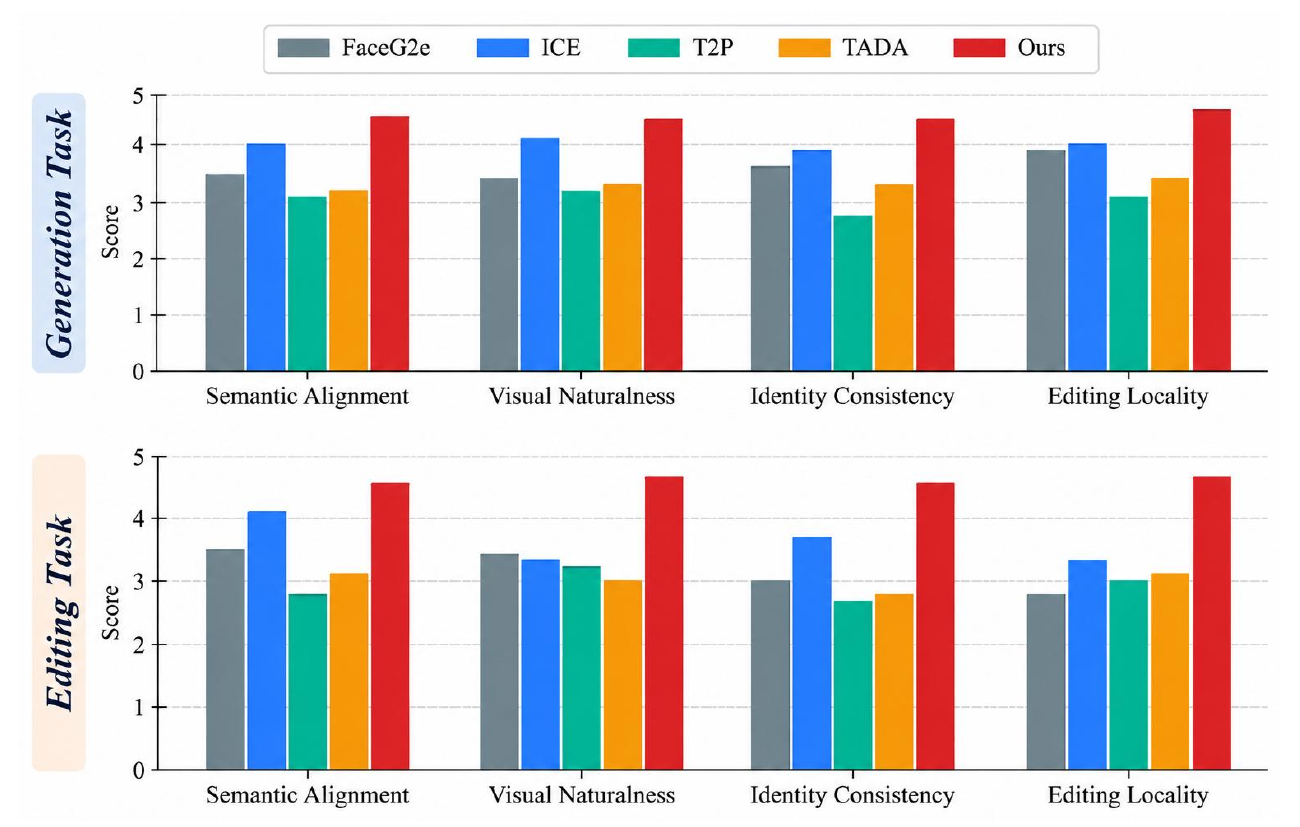}
 \vspace{-3mm}
 \caption{User study for 3D facial generation and editing result.}
 \vspace{-3mm}
 \label{userstudy}
\end{figure}

User evaluation is critical for assessing both generation quality and interaction performance of text-driven 3D face editing models. For a comprehensive comparative analysis, 32 participants were recruited to subjectively evaluate the outputs of T2P~\cite{T2P}, TADA~\cite{TADA}, FaceG2e~\cite{FaceG2E}, ICE~\cite{ICE}, and our RAGMesh. Participants were instructed to complete generation and editing tasks using each method, and rated the results against four evaluation criteria: (1) semantic alignment, (2) visual naturalness, (3) identity consistency, and (4) editing locality. As illustrated in Figure~\ref{userstudy}, RAGMesh achieves consistently higher subjective scores across all metrics, which validates its superior semantic controllability, geometric naturalness, and identity preservation, as well as a more intuitive and fluent user interaction experience. (\textit{Refer to SM for user study questionnaire}).

\subsection{Feature-Level Analysis of RAG}
\begin{figure}[t]
 \centering
 \includegraphics[width=1\columnwidth]{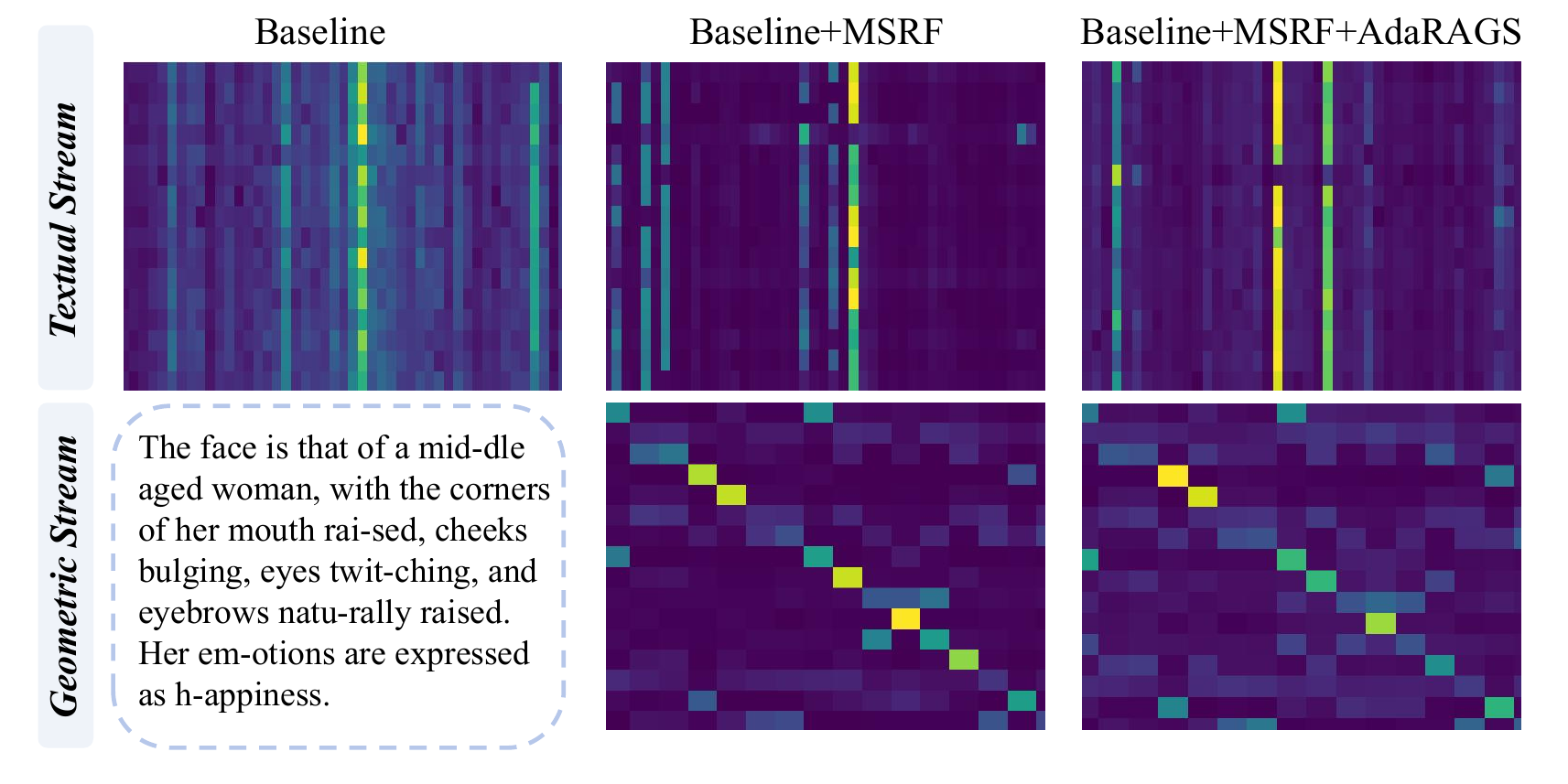}
 \vspace{-3mm}
 \caption{Visualization of intermediate feature representations as the MSRF and AdaRAGS modules are introduced into the model.}
 \label{feature}
 \vspace{3mm}
 \centering
 \includegraphics[width=1\columnwidth]{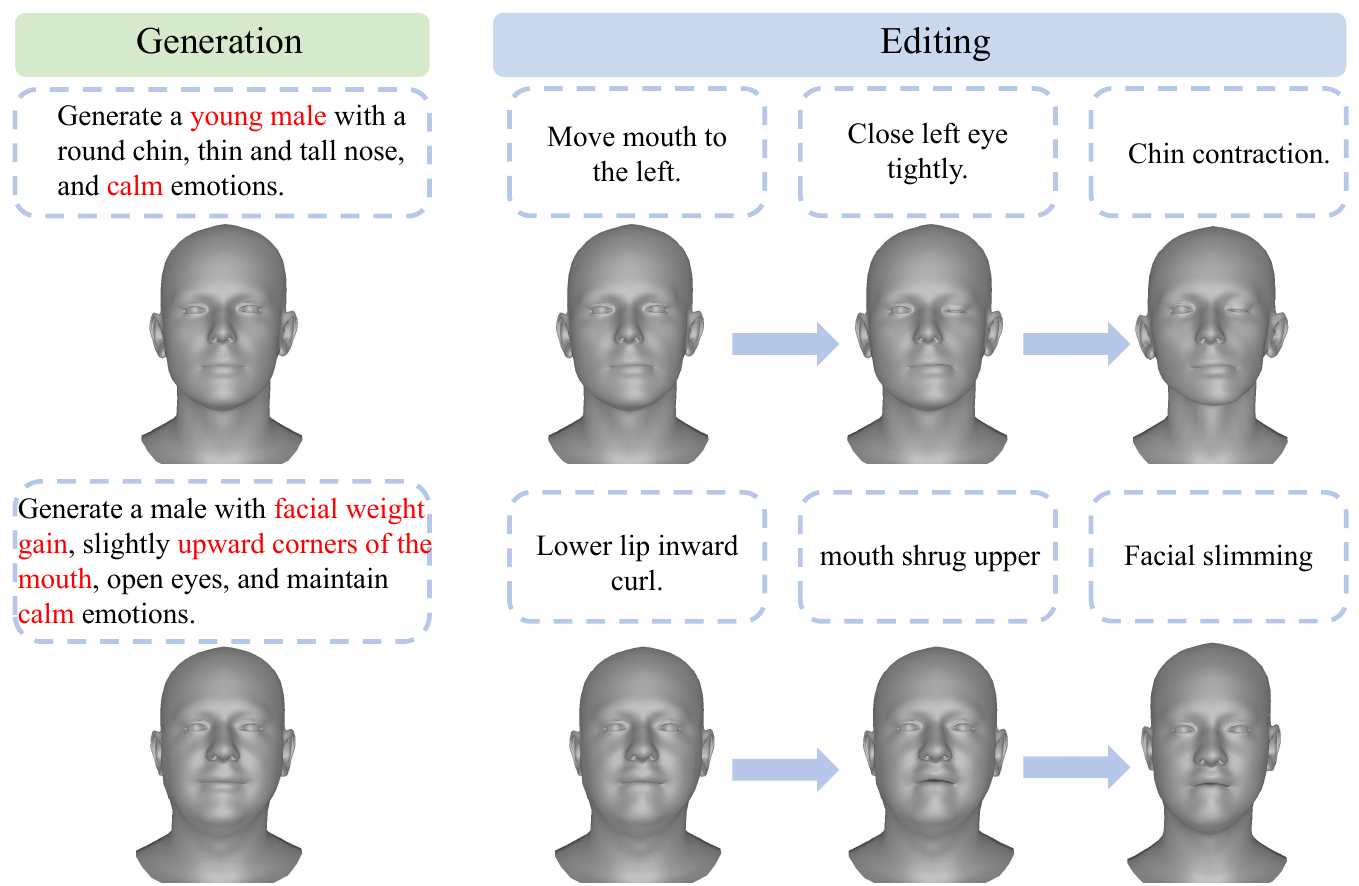}
 \vspace{-3mm}
 \caption{Progressive text-guided 3D face editing. Our method sequentially applies multiple local editing instructions to the generated faces while preserving identity consistency and overall facial geometry.}
 \vspace{-3mm}
 \label{edit}
\end{figure}

Figure~\ref{feature} visualizes text and geometric feature representations with MSRF and AdaRAGS.
The baseline focuses on dominant tokens in long text descriptions, which correspond to regions with large geometric variations. While capturing global deformations, it neglects semantically critical but subtle local details. By contrast, MSRF introduces extra geometric priors and strengthens the model’s representation capability. Furthermore, AdaRAGS provides region-level guidance from locally retrieved meshes, which reallocates attention in the feature space and enables the model to better capture fine-grained details ignored by the baseline.

\subsection{Progressive Continuous Editing}
For continuous facial editing, we first employ the generation model to produce an initial facial mesh. Since both the generation and editing models operate in the FLAME geometric space, the generated mesh is directly used as the input for subsequent editing. Given a new editing instruction in the text domain, RAGMesh retrieves the most relevant facial geometries from the database as geometric priors to guide the editing process. As illustrated in Figure~\ref{edit}, the retrieval-conditioned editing pipeline enables progressive and consistent facial modifications. Throughout the editing process, the identity of the character is well preserved, while fine-grained and region-specific facial attributes can be precisely controlled.

\subsection{Discussion and Limitation}
RAGMesh adopts a geometry-centric design that does not explicitly model texture or appearance. This design isolates fine-grained geometric controllability and text–geometry alignment, while maintaining compatibility with existing rendering pipelines, serving as a controllable geometric backbone that complements full appearance-generation systems. However, the separate training of generation and editing tasks restricts parameter sharing and compromises cross-task consistency. Consequently, generated meshes may not always fully adhere to fine-grained textual descriptions, necessitating iterative refinement via the editing module (\textit{see SM for failure cases}). A promising direction is to unify generation and editing in a single architecture with shared representations to enable mutual supervision and enhance controllability and fidelity.


\section{Conclusion}
In this paper, we first construct FaME-G2E, a large-scale multimodal 3D facial dataset containing text--mesh pairs for generation and text--blendshape pairs for localized editing. Based on this dataset, we propose RAGMesh, a retrieval-augmented framework that leverages geometric priors for text-driven 3D facial generation and editing. Specifically, MSRF retrieves semantically relevant reference meshes to enhance geometric fidelity, while AdaRAGS leverages region-aware supervision for fine-grained text-to-geometry alignment. Extensive experiments demonstrate that RAGMesh achieves superior performance over state-of-the-art methods in semantic consistency, geometric fidelity, diversity, editing locality, and inference efficiency. Future work will explore more diverse facial assets, unified generation-editing models, and extensions to broader 3D domains and multimodal interactions.

\section*{Acknowledgments}
{\color{black}\textbf{Ethics Statement.}} This work involved human subjects in its research. Approval of all ethical and experimental procedures and protocols was granted by Biological and Medical Ethics Committee of Beihang University (IF PROVIDED under Application N0.BM20230165, and performed in line with the Approval Letter from the Biological and Medical Ethics Committee of Beihang University)


\bibliography{template}
\bibliographystyle{IEEEtran}

\end{document}